# Application of Faster R-CNN Model on Human Running Pattern Recognitions

Kairan Yang, Feng Geng
Lexington High School, KTByte Computer Science Academy

Abstract
The advance algorithms like Faster Regional Convolutional Neural Network (Faster R-CNN) models are suitable to identify classified moving objects, due to the efficiency in learning the training dataset superior than ordinary CNN algorithms and the higher accuracy of labeling correct classes in the validation and testing dataset. This research examined and compared the three R-CNN type algorithms in object recognitions to show the superior efficiency and accuracy of Faster R-CNN model on classifying human running patterns. Then it described the effect of Faster R-CNN in detecting different types of running patterns exhibited by a single individual or multiple individuals by conducting a dataset fitting experiment. In this study, the Faster R-CNN algorithm is implemented directly from the version released by Ross Girshick.

1. Introduction:
   Object detection is a subfield in machine learning, and this technology is vastly applied to real-life events for various purposes. In particular, the detection of moving objects exhibiting certain behaviors or patterns is important to the security field and public safety field. Our project is inspired by the public safety concern. Specifically we want to identify potential school emergencies and dangers from recognizing motion patterns of the students. There are many algorithms aiming at improving the accuracy and efficiency of the visual recognitions and classifications. Examining those algorithms, the most suitable algorithms for this particular set of constraints are the Regional Convolutional Neural Networks (R-CNNs), primarily contributed by Ross Girshick from University of California, Berkeley from 2014 to June 2015. This algorithm family consists of three subsequent algorithms, namely R-CNN, Fast R-CNN and Faster R-CNN [1],[4],[5].

   This project is an application research based on the computer vision algorithm Faster R-CNN devised by Shaoqing Ren and Ross Girshick, et al [5]. To separate the "emergency" case from the ordinary setup, we classified three different types of human motions: walking, jogging, and panically running (aka escaping). The goal is to train the model to be able to recognize single or multiple individuals exhibiting panically running pattern from the data set comprised of all three types of human motions described above.

In short, this research aims at using Faster R-CNN to fit a particular dataset of our interest.

The paper is divided into three major sections. The first section describes Faster R-CNN and its predecessors: R-CNN and Fast R-CNN, in order to show the efficiency of using Faster R-CNN rather than other methods or the preceded methods. The second section describes the research itself from the process of crafting the training, cross-validation and testing dataset to the training process, and to the analysis of the experimental results of the dataset. To achieve this, we incorporated the Faster R-CNN source code [8] into the GPU server designated for this training. The third section is the conclusion drawn by the experiment result, and the talk of further applications of the Faster R-CNN algorithm and this specific dataset on computer vision and its prospective contribution on the campus safety issues.

2. Methodology:

Algorithms of object detection are constantly evolving. In 2014 Girshick et al. from UC Berkeley devised and proposed the R-CNN algorithm, which stands for regionalizing CNN features[1], extracting features based on distinct regions in the input images rather than scanning square pixel matrices on the image as a whole. In other word, R-CNN preprocessed feature extraction by compartmenting the image into various sub-images, then it performed feature extraction separately on each of the sub-image. Then, the R-CNN model using class-specific linear SVM to classify the result of the convoluted features. R-CNN improved its accuracy of classification over traditional fitting algorithms, achieving a mean average precision (mAP) of 53.7% on the dataset of PASCAL VOC 2010 and 34.3% on ILSVRC2013 detection dataset[1],[6],[9], outperforming the regular methods such as deformable part model (33.4% on PASCAL VOC 2010 [2], [6]) and Overfeat (24.3% on ILSVRC2013 [3], [9]).

However, the expanse of computational space and time for performing ConvNet on each of the region proposals and regressing the features using linear SVMs [4] had greatly impeded the efficiency of R-CNN algorithm. According to Ross Girshick in his article *Fast R-CNN*, "Detection with VGG16 takes 47s / image (on a GPU)." [4] To overcome this drawback, Girshick devised Fast R-CNN network. The architecture of Fast R-CNN network consists the RoI(region of interest) pooling layer which is an improved version of distinct regions of the input image, defined by four parameters *(r ,c, h, w)*, where *(r , c)* is the Cartesian coordinate of the top-left corner of the feature map and *(h, w)* is the bottom-right. RoI pooling layers are pooled into fixed-size feature map to subsequently input into the dense layers and then followed the regular CNN classification process.

Using Fast R-CNN does increase the mAP, performing 66.1% on the dataset VOC 2010 [4],[6], which is 12.4% greater than the accuracy of regular R-CNN. However, both the training and testing efficiency have been significantly boosted. According to Girshick, "For VGG16, Fast R-CNN processes images 146× faster than R-CNN without truncated SVD and 213× faster with it." [4] Also, on the testing result of VOC07 dataset, the test rate increases to 0.32 seconds per image.

In June 2015, Shaoqing Ren et al. devised Faster R-CNN based on the Regional Proposal Network (RPN) to further increases the efficiency of processing images for object recognition [5]. The RPN revolutionizes the method of finding regional proposals from selective search that R-CNN and Fast R-CNN used to a convolutional neural network that "simultaneously predicts object bounds and objectness scores at each position." [5] Then, Ren et al. combined the RPN network of generating features and the old Fast R-CNN architecture together, still using the RoI pooling layers to further process and classify the objects within the image. Another aspect of Faster R-CNN is its ability to predict multiple proposals simultaneously. The architecture of RPN has sliding window implementation. Each sliding window location on the feature map could afford a number of regional proposals, denoted $k$, based on the scale and aspect ratio variables -- which are hyperparameters in the training process. In the center of the sliding windows there are $k$ anchor boxes, corresponding to the $k$ regional proposals that will be produced. Ren et al. utilizes these anchors in a stacking fashion, known as "multi-scale anchors" [5], essentially a pyramid of anchors to improve the efficiency of generating regional proposals. According to Ren et al., the optimal anchor setting in terms of mAP performance, is 3 scales and 3 ratios, in total to make $k$ equal to 9. This setting led Faster R-CNN perform 69.9% on the PASCAL VOC 2007 dataset, ranked the best among the 4 combinations of scales=1 or 3 and ratios=1 or 3.

Ren et al. compared the performance of Fast R-CNN and Faster R-CNN on the Microsoft COCO (MS COCO) object detection dataset [7]. MS COCO dataset has two metrics to evaluate mAP: mAP@.5 and mAP@[.5, .95] [5], but we decide to use the mAP@.5 metric since it is identical to the mAP metric used in the VOC datasets, to make the comparison more direct. For the MS COCO dataset, the training accuracy of Faster R-CNN using RPN and 300 regional proposals in terms of mAP@.5 metric is 42.1% [5], while the training mAP@.5 of selective search Fast R-CNN with 3000 regional proposals is only 35.9% [4], [5]. The difference is not significant in terms of the 6.2% outperformance compared to the 12.4% mAP difference between Fast R-CNN and regular R-CNN. However, the test rate does increase. According to Ren et al., on the MS COCO dataset the reported test rate is "about 200ms per image" [5] (200 ms equal to 0.2 seconds). Comparing those three algorithms described above, we finally decide to use

Faster R-CNN -- the most efficient and most accurate method among the three based on the data provided by [1], [4], and [5].

3. Data Set
3.1 Data Set Introduction and Data Collection

The dataset of this research is collected as a series of m4v 1920x1080p iPhone 6 camera shot video clips. During the course of data collection phase, we collected about 12 minutes of video clips about different people walking, jogging and panically running. Each individual video clip is about 20 seconds long, and each video is only recording one type of the three motions described above. Therefore, 36 video clips were recorded in total. There might be a single or multiple subjects appeared in any video clips. For our dataset, 6 out of the 36 video clips are having multiple figures shown and exhibited uniform type of motions, *i.e.* multiple subjects performed only one motion type (walking, jogging or panically running) throughout the 20 second interval. And the other 30 video clips are about a single subject performing one motion type over the video clip's duration. No individual subject is tested repeatedly except those who were asked to perform in a group. Surprisingly, we did not focus on maintaining a school like background when recording the videos, because of the Faster R-CNN algorithm is translation-independent[5], which means the background setting does not have observable impact on the motion recognition process in later training.

To collect the data, we recruited and relied exclusively on the volunteers who performed each of the moving patterns for 20 seconds, in total contributing for one minute of raw data videos. We noticed and explained to every single volunteer about the purpose and content of this research project, and their duties during the recording process. We asked each of them to perform all three types of motion to ensure the data is categorized. The recording events are held in separate time period, one on August 10, 2018 at the Mudge House of Carnegie Mellon University and another on August 30, 2018 at the front courtyard our friend's house in Sharon, MA. Only 2 volunteers participated in the first video recording event, resulted in approximately 3 minute dataset length (9 video clips, 6 individual, 3 group). All volunteers are CMU enrolled undergraduate students. In comparison, 7 volunteers were recruited in the second video recording event. Collectively about 9 minutes of raw data were gained, resulted in 27 video clips in which 9 are group clips and 18 are single clips. The group clips contained 1 set of 2-person motion clip and 1 set of 3-person motion clip to diversify the dataset. (a set of clips is defined as the collection of three motion clips of one individual or group encompassed walking, jogging and escaping motions) All volunteers in the Sharon recording event are adults aged from

30 to 40, older than the volunteers from the first event but geometrically their body shapes and motions are not significantly different than those of teenage students.

|  | Mudge Recordings | Sharon Recordings |
|---|---|---|
| Single (one person) clips | 6 clips, 122 seconds | 18 clips, 364 seconds |
| Multiple (many people) clips | 3 clips, 67 seconds | 9 clips, 185 seconds |

Table 1: Categories of video clips based on the number of volunteers appeared in the camera (single/multiple).

|  | Mudge Recordings | Sharon Recordings |
|---|---|---|
| Walking clips | 3 clips, 63 seconds | 9 clips, 180 seconds |
| Jogging clips | 3 clips, 65 seconds | 9 clips, 182 seconds |
| Escaping clips | 3 clips, 61 seconds | 9 clips, 187 seconds |

Table 2: Categories of video clips based on the type of motions each clip exhibited (walking/jogging/escaping).

3.2 Data Set Compilation From Raw Video Format to JPEG Format

In order to convert the raw data video clips to image inputs. We decide to import the video clips from iPhone storage to KMPlayer, a software able to extract image frames from mp4/m4a format videos. Because the labeling helper class in the training program specified JPEG format inputs, we specified the converted type to be JPEG image files.

We extracted 200-300 images from each clip to ensure the dataset to be large enough to avoid high bias underfitting, but not too large to make the manual labeling process be exhausted. For the 36 video clips we recorded from the volunteers, 11 video clips (all from the second recording event that held on the evening) exhibited poor lighting conditions which could result in indistinguishable blur between the background and the person. We eliminated one more to ensure each motion type held the same number of video clips. For the 24 videos left, we extract 200 or 300 images uniformly across the frames of the video clips, resulted in 6700 JPEG image files being extracted. However, due to the limitation of iPhone 6 camera during the recording event, a small portion of the images exhibited some degree of blur and/or body parts not captured by the camera.

Eliminating those images, there are 4,586 images valid to be input data for the training dataset. We randomly choose 3,000 images to be the training set (65.4%), 793 images to be the cross-validation set (17.3%) and 793 images to be the testing set (17.3%).

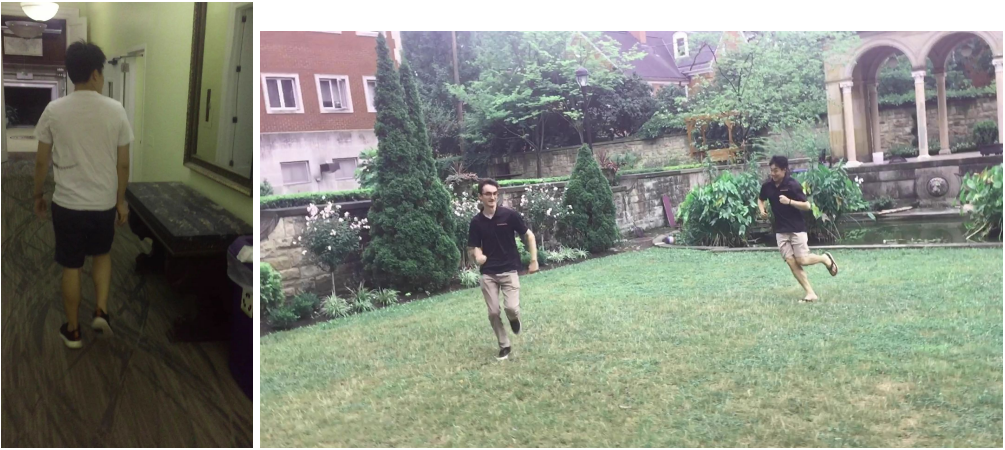

Figure 1: two valid training images from the research dataset. The left one is an example of single walking category while the right one belongs to the multiple escaping category.

| Image Type | Single | Multiple | Total |
| --- | --- | --- | --- |
| Walking (a) | 1,031 | 553 | 1,584 |
| Jogging (b) | 1,018 | 498 | 1,516 |
| Escaping (c) | 966 | 520 | 1,486 |
| Total | 3,015 | 1,571 | 4,586 |

Table 3: The two-way frequency table of dataset image categories

3.3 Data Preprocessing

The purpose of preprocessing the dataset is to convert image files to xml files in order to feed the training neural network. To achieve this, we devised a helper package written in python to crop the images and label the type of motions, and to convert the data from 4,586 discrete image files to three single xml matrix files where each row represented one image example and stored all vital information of the image to be fed into the neural network. The files of the helper package were uploaded to the GitHub repository *https://github.com/kairanYang/runningPatternRecognition*. [11]

To categorize the three different motion types, we labeled three categorical traits for the images from the dataset, namely *a*, *b*, and *c* corresponding to the motion types of walking, jogging and escaping respectively. We label everything in the research dataset by hand. Specifically, we label the images using *mark.py* [11] in the helper package. *mark.py* writes a txt intermediate file for each image file that contains the label (97=*a*, 98=*b*, 99=*c*), the width and length of the image expressed in numbers of pixels (*(1080, 1920)* throughout the dataset) , the cartesian coordinate of the top-left and bottom-right corner of the label crop (usually an rectangle enclosed one person or more people exhibiting motions in the image). After all images are processed into txt files, these information contained in the txt files will move to *converttxt2voc.py* [11] to be compiled as xml files: one for training set, one for cross-validation set and one for testing set.

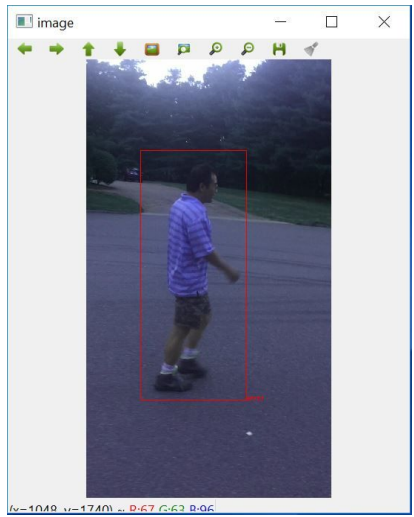

Figure 2: Executing mark.py to crop the image from the dataset.

4. Training Process and Testing Result

4.1 Training Process

We incorporated the process of generating automatic regional proposals in the training, namely, the RPN network. Since Faster R-CNN spend only about 200ms to process a single image, and the accuracy of the regional proposals to cover the region of interest, *i.e.* "the person", is enough to satisfy the constraint of this research. This dataset does not contain too much obstacles, noises or blurs, so it is a relatively easy dataset to train compared to the MS COCO dataset [7], PASCAL 2007 [10] or PASCAL 2010 [6]. Also, the mAP performance is 42.1% on the MS COCO dataset [5], [7], outperforming every other R-CNN type algorithms [5], presumably the high performance of Faster R-CNN will be repeated on this image dataset. To maximize the accuracy, we set the scale factor and ratio factor of 3, resulted *k*, the maximum regional proposal number for each sliding window in the RPN, to be 9. With these concerns and confidences, we trained the 3000 image training dataset on September 14, 2018. The training process is entirely based on Girshick's Faster R-CNN python code [8], which has been released to public use since 2015. The training took 6 hours and 33 minutes to complete on the Linux Ubuntu 16.04 virtual machine. The Faster R-CNN model fitted 99.8% on the training dataset. From the training result, we are confident that the model would fit above 99% in the testing dataset.

4.2 Testing Result

The model was then validated on the separate 793-image testing dataset. The accuracy of Faster R-CNN model on this dataset is 98.5%, identified the correct motion pattern of 781 images out of the 793 images total. For the 12 misidentified images, 10 of them are mislabeled into other categories and 5 of them identified incorrect region of interest, 3 images exhibited both problems This result could be interpreted as about 98.5% of the situations, the correct motion pattern would be identified correctly. About 1.5% of the situations there could be incorrect classifications or incorrect region of interest identified, which is higher than we expected before the validation process. Therefore, we are confident that the training model is subjected to some degree of high variance. We can also calculate the F1 score for this dataset. As described above, for the 793 images in the testing dataset, 781 of them are true positive, 10 mislabeled images are false positive, and 5 mislocated images are false negative. Then, the precision and recall are 0.9874 and 0.9936, respectively. The F1 score is therefore 0.9780. For the most concerned escaping or $c$ category, the categorical fitness specifically for group $c$ is 97.7%, lower than the average fitness.

| Label Type | Correctly Identified | Total Labels | Percentage |
| --- | --- | --- | --- |
| Walking (a) | 263 | 266 | 98.9% |
| Jogging (b) | 260 | 263 | 98.9% |
| Escaping (c) | 258 | 264 | 97.7% |
| Total | 781 | 793 | 98.5% |

Table 4: Label-wise or motion-wise fitness table of the testing dataset.

## 5. Conclusion

This research study does prove that the Faster R-CNN is a powerful, efficient and accurate human running pattern detector model in highly simulated realistic environments. However, according to the result presented in section 4.2, as the motion become more complex, the accuracy of successful detection decreases. This might be attributed to the fact that faster moving speed causes the images to blur, weakening the edge between human body and the background. We also conjectured that the accuracy may decrease due to the generalization of datasets, because our model has shown clear overfitting problem by comparing the accuracies of the model performing on the training and testing sets.

## Acknowledgements

I thank Mr.Feng Geng, who is the mentor of this research project. Mr.Geng has given me enormous amount of support academically and mentally. We both have many wonderful discussions about the experimental design, coding, and writing formal academic papers. I thank to Mr.Benjamin Po from KTByte Academy to give me clarifications when explaining hard concepts. I also thank to Professor Gordon Weinberg of Carnegie Mellon University, who approved and supported this project to be conducted in CMU campus. Last but not the least, thank for Tyler Stoner, Roy Xu, Xiufeng Wu, and all volunteers in the video clip recording events.